\documentclass{article} 
\usepackage{iclr2024_conference,times}

\usepackage{amsmath,amsfonts,bm}

\def\eqref#1{equation~\ref{#1}}

\def\1{\bm{1}}

\DeclareMathAlphabet{\mathsfit}{\encodingdefault}{\sfdefault}{m}{sl}
\SetMathAlphabet{\mathsfit}{bold}{\encodingdefault}{\sfdefault}{bx}{n}

\usepackage{hyperref}
\usepackage{url}
\usepackage{booktabs}
\usepackage{multirow}
\usepackage{colortbl}
\usepackage[normalem]{ulem}

\usepackage{graphicx} 

\title{Extending Multi-modal Contrastive Representations}

\author{Zehan Wang\thanks{Equal contribution. Contact: \{wangzehan01, ziangzhang\}@zju.edu.cn} \\
Zhejiang University
\And
Ziang Zhang\footnotemark[1] \\
Zhejiang University
\And
Luping Liu \\
Zhejiang University
\And
Yang Zhao \\
ByteDance
\And
Haifeng Huang \\
Zhejiang University
\And
Tao Jin \\
Zhejiang University
\And
Zhou Zhao \\
Zhejiang University
}

\iclrfinalcopy
\begin{document}

\maketitle

\begin{abstract}
Multi-modal contrastive representation (MCR) of more than three modalities is critical in multi-modal learning. Although recent methods showcase impressive achievements, the high dependence on large-scale, high-quality paired data and the expensive training costs limit their further development. Inspired by recent C-MCR, this paper proposes \textbf{Ex}tending \textbf{M}ultimodal \textbf{C}ontrastive \textbf{R}epresentation (Ex-MCR), a training-efficient and paired-data-free method to flexibly learn unified contrastive representation space for more than three modalities by integrating the knowledge of existing MCR spaces. Specifically, Ex-MCR aligns multiple existing MCRs into the same based MCR, which can effectively preserve the original semantic alignment of the based MCR. Besides, we comprehensively enhance the entire learning pipeline for aligning MCR spaces from the perspectives of training data, architecture, and learning objectives.
With the preserved original modality alignment and the enhanced space alignment, Ex-MCR shows superior representation learning performance and excellent modality extensibility. To demonstrate the effectiveness of Ex-MCR, we align the MCR spaces of CLAP (audio-text) and ULIP (3D-vision) into the CLIP (vision-text), leveraging the overlapping text and image modality, respectively. Remarkably, without using any paired data, Ex-MCR learns a 3D-image-text-audio unified contrastive representation, and it achieves state-of-the-art performance on audio-visual, 3D-image, audio-text, visual-text retrieval, and 3D object classification tasks. More importantly, extensive qualitative results further demonstrate the emergent semantic alignment between the extended modalities (e.g., audio and 3D), which highlights the great potential of modality extensibility. Our code is available at \url{https://github.com/MCR-PEFT/Ex-MCR}.
\end{abstract}

\section{Introduction}

Multi-modal Contrastive Representation (MCR) learning endeavors to align inputs from diverse modalities within a shared representation space. Recently, the high-quality contrastive representations of more than three modalities attract increasing attention~\citep{girdhar2023imagebind, guzhov2022audioclip, xue2023ulip, xue2023ulip2, liu2023openshape, hegde2023clip, guo2023pointbind}, and play a fundamental role in many application scenarios of multi-modal understanding~\citep{su2023pandagpt, zhang2023video, zhao2023bubogpt, wang2023chat, han2023imagebind} and generation~\citep{tang2023anytoany, liu2023audioldm, ramesh2022hierarchical, rombach2022high, gafni2022make, huang2023make}. Despite the achievements of multi-modal contrastive learning, its broader and more flexible application is still constrained by the high dependence on large-scale, high-quality paired data and extremely costly training resources. 

Recently, \cite{wang2023connecting} introduces a novel training-efficient method, called C-MCR, for learning contrastive representations between modalities that lack paired data by mining knowledge from existing MCR spaces. It connects two pre-trained MCRs onto a new shared space via overlapping modalities. Since the modalities of each MCR are intrinsically aligned, the connection learned from overlapping modalities can also be transferred to non-overlapping modalities.
Experimentally, without using image-audio and 3D-text data pairs, C-MCR demonstrates advanced performance in image-audio and 3D-text downstream tasks.

Despite the remarkable flexibility and performance of C-MCR, its broader applications are hindered by a critical limitation: C-MCR mainly focuses on learning a new space for the two non-overlapping modalities, while the original modality alignments in powerful pre-trained MCRs are forgotten. As a result of the decline of original alignment, C-MCR faces challenges in concurrently establishing connections among three or more MCRs. Therefore, C-MCR can not be used to flexibly learn a shared contrastive representation space for more than three modalities.

This paper introduces \textbf{Ex}tending \textbf{M}ulti-modal \textbf{C}ontrastive \textbf{R}epresentations (Ex-MCR), a novel training-efficient and paired-data-free unified representation learning method with excellent modality extensibility. Ex-MCR better preserves the alignment within the original pre-trained MCR space and enhances the overall learning pipeline to align different MCR spaces more robustly. Specifically, the two important designs of Ex-MCR are discussed in detail below:

Firstly, we extend one MCR space (called leaf-MCR) into another fixed MCR space (called base-MCR) rather than connecting two MCR spaces to a new space. Such a simple yet effective approach maximizes the preservation of modality alignment within the base MCR, demonstrating great potential for integrating multiple MCRs.

Secondly, we enhance the whole learning process to promote stronger alignment across different MCRs. Specifically: 1) From the training data perspective, we extract various modality-centric pseudo data pairs, aiming to alleviate the semantic bias of pseudo pairs in~\cite{wang2023connecting} and reflect MCR space more comprehensively. 2) From the architecture perspective, we propose a decoupled projector, which reduces interference among different optimization objectives. We further find that a simple linear mapping is more effective for learning to eliminate modality gaps within MCRs. 3) From the learning objective perspective, we employ a dense contrastive loss on pseudo-pairs between all possible modalities pairs, further enhancing the learned alignment's stability.


Utilizing Ex-MCR, we can flexibly align multiple pre-trained leaf-MCR spaces onto a common base-MCR space without any paired data and with exceptionally low training costs. To evaluate the effectiveness of our Ex-MCR, we try to extend the ULIP (3D-image) and CLAP (audio-text) onto CLIP (image-text) via the overlapping image and text modality, respectively, which derive unified and high-quality audio-image-text-3D representations. Without using any paired data, Ex-MCR attains state-of-the-art performance results across various zero-shot tasks, including audio-visual, 3D-image, audio-text, visual-text retrieval, and 3D object classification. More importantly, semantic alignment is also observed between extended modalities (e.g., audio-3D), which highlights the potential of Ex-MCR in modality extensibility.

Our contributions can be summarized as three-fold:

(1) We propose \textbf{Ex}tending \textbf{M}ulti-modal \textbf{C}ontrastive \textbf{R}epresentations (Ex-MCR), a novel a training-efficient and paired-data-free representation learning method for more than three modalities.

(2) We comprehensively enhance the entire learning pipeline for aligning MCR spaces from the perspectives of training data, architecture, and learning objectives. These novel designs offer valuable insights about effectively integrating knowledge within existing MCRs.

(3) We obtain high-quality unified audio-image-text-3D representations using Ex-MCR, which exhibits advanced performance on a series of tasks and excellent modality scalability. Besides, we also conduct detailed ablation studies to verify the effectiveness of each proposed component.


\section{Related Works}
\subsection{Multi-Modal Contrastive Representations}

Multi-modal Contrastive Representations (MCR) learning aims to acquire semantically aligned cross-modal representations by pretraining the model on large-scale paired data. These aligned representations play a pivotal role in downstream comprehension and generation tasks. Inspired by the success of CLIP~\citep{radford2021learning}, many works try to learning contrative representations for two modalities~\citep{radford2021learning, li2022blip, li2021align, gan2022vision, xu2021videoclip}. CLIP~\citep{radford2021learning} and ALIGN~\citep{jia2021scaling} learn shared vision-text representations from million-level image-text pairs. CLAP~\citep{elizalde2023clap, wu2023large} learns the audio-text representation, and CAV-MAE~\citep{gong2022contrastive} focus on acquiring shared audio-visual feature space. C-MCR~\citep{wang2023connecting} focuses on learning new representation space by connecting the pre-trained spaces through overlapping modality. 

Apart from aligning two modalities, shared representations for more than three modalities attract increasing attention. AudioCLIP~\citep{guzhov2022audioclip} and WAV2CLIP~\citep{wu2022wav2clip} train an audio encoder aligned with CLIP using audio-text-image triplets data. ULIP~\citep{xue2023ulip, xue2023ulip2} and openshape~\citep{liu2023openshape} construct 3D-image-text triplets data through rendering 3D mesh into 2D images and captioning images for textual description, thereby learning a corresponding 3D encoder for image-text MCR space. Furthermore, Imagebind~\citep{han2023imagebind} exclusively utilizes data pairs between various modalities and images to expand CLIP with multiple modal alignment encoders. 

However, these methods heavily rely on large-scale, high-quality paired data collected from the internet or generated automatically and exceptionally high computational resources. Due to the lack of high-quality paired data for more modal combinations, such as audio-visual and text-3D, the extensibility of representation learning is notably constrained. Furthermore, the exceedingly high computational costs also diminish the flexibility of MCR learning.

\subsection{Audio-Visual and 3D-Text Learning}
Audio-vision and 3D-text learning have significant applications in multi-modal recognition~\citep{gemmeke2017audio, chen2020vggsound, chang2015shapenet, dai2017scannet}, localization~\citep{chen2020scanrefer, achlioptas2020referit3d, zhao20213dvg, zhao2018sound, mo2022closer, chen2021localizing}, question-answer~\citep{wang2023chat, zhao2023bubogpt, azuma2022scanqa, lin2023vision}, and generation~\citep{ruan2023mm, poole2022dreamfusion, lin2023magic3d}. They also play important roles in robot-related tasks such as human-machine interaction and synthetical information obtaining in complex environments~\citep{peng2023kosmos, huang2023language}.

However, audio-visual datasets~\citep{gemmeke2017audio, chen2020vggsound} often suffer from substantial noise due to soundless objects and invisible sounds. Additionally, paired 3D-text data~\citep{chang2015shapenet} is scarce and expensive to collect. The scarcity of large-scale datasets hampers the further advancement of 3D-text and audio-vision contrastive representations. Previous methods, such as AudioCLIP~\citep{guzhov2022audioclip} and ULIP~\citep{xue2023ulip, xue2023ulip2}, mainly focus on automatically collecting or generating more paired data, but they are still limited by the relatively low quality of the training datasets. Our approach overcomes the reliance on paired data, achieving superior performance in audio-vision and 3D-text retrieval without using any audio-vision or 3D-text data.

\section{Extending Multi-modal Contrastive Learning}

\subsection{Extending Rather Than Connecting}

Given two pre-trained MCR spaces on modalities $(\mathcal{A}, \mathcal{B})$ and $(\mathcal{B}, \mathcal{C})$, C-MCR~\citep{wang2023connecting} employs two projectors to map them into a new shared space, where the alignment of different MCRs can be learned from overlapping modality $\mathcal{B}$. Since each pre-trained MCR intrinsically contains the alignment of $(\mathcal{A}, \mathcal{B})$ and $(\mathcal{B}, \mathcal{C})$, the alignment learned from overlapping modality theoretically can be transferred to the non-overlapping modalities. Specifically, the embeddings from overlapping modality $\mathcal{B}$ but different MCR are aligned via an InfoNCE loss in the new space. Besides, C-MCR retrieves pseudo $(\mathcal{A}, \mathcal{C})$ pairs using the same data of $\mathcal{B}$ and these pseudo-pairs are also aligned for a more comprehensive inter-MCR alignment. Moreover, C-MCR employs L2 loss between the embeddings from the same MCR space but different modalities to close the modality gap~\citep{liang2022mind}, significantly enhancing the transferability of learned inter-MCR alignment. C-MCR has remarkable flexibility and versatility since learning a novel C-MCR space requires two learnable MLPs and unpaired unimodal data.

However, C-MCR mainly focuses on learning a new space for the two non-overlapping modalities ($\mathcal{A}$, $\mathcal{C}$), while the original modality alignment ($\mathcal{A}$, $\mathcal{B}$) and ($\mathcal{B}$, $\mathcal{C}$) in powerful pre-trained MCRs are forgotten. As a result of the decline of original alignment, C-MCR faces challenges in concurrently establishing connections among three or more MCRs. Therefore, C-MCR can not be used to flexibly learn a shared contrastive representation space for more than three modalities.

To achieve a training-efficient and paired-data-free unified contrastive representation method, we propose to extend one MCR to another rather than connect the two MCRs to a new space. 
Considering the two MCR spaces on modalities $(\mathcal{A}, \mathcal{B})$ and $(\mathcal{B}, \mathcal{C})$, Ex-MCR chooses one as the base-MCR $(\mathcal{A}, \mathcal{B})$, and the other as the leaf-MCR $(\mathcal{B}, \mathcal{C})$. In the ``Extended" scheme, the base-MCR space is frozen, and we only train one projector to map leaf-MCR to base-MCR utilizing the overlapping modalities $\mathcal{B}$. Specifically, we employ the native pairs of $\mathcal{B}$ and pseudo pairs generated by $\mathcal{B}$ to learn aligning leaf-MCR to base-MCR via InfoNCE loss. Simultaneously, we employ the L2 loss to bridge the modality gap between $(\mathcal{B}, \mathcal{C})$ modalities of leaf-MCR, thereby facilitating more transferable alignments between the MCR spaces.

In contrast to C-MCR, Ex-MCR can conveniently expand more MCR spaces. Benefiting from efficient training and no need to pair data, we can flexibly align multiple leaf-MCR spaces to the same base-MCR space. In addition to explicitly establishing modality alignment within leaf-MCR and base-MCR, semantic alignment also emerges among extended modalities. Ex-MCR leverages the pivotal role of the base-MCR, employing it as a bridge for achieving semantic alignment among modalities in multiple leaf-MCR spaces.

\begin{figure*}
	\centering
	\includegraphics[width=1\linewidth]{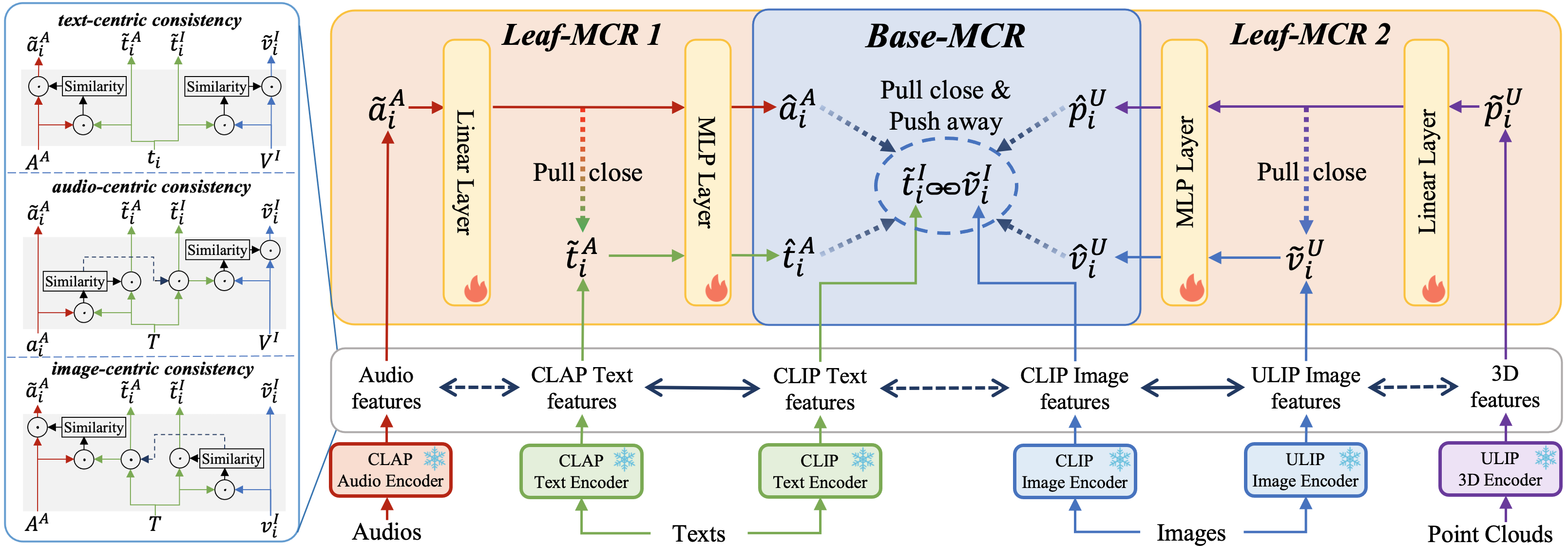}
	\caption{\textbf{The pipeline of extending leaf-MCRs (CLAP, ULIP) to base-MCR (CLIP).} For aligning CLAP to CLIP, we take audio, text, and image as input and encode them individually with frozen encoders in CLAP and CLIP. As shown in the left subfigure, we iteratively take the three kinds of modalities as query to generate pseudo-pairs. For aligning ULIP to CLIP, we take a symmetrical approach. When inferencing, audio and 3D inputs are inputted to the CLAP audio encoder and ULIP 3D encoder, then mapped into the CLIP MCR space via the corresponding projectors. Texts and images are encoded by the CLIP text encoder and image encoder.}
 
        \label{fig:pipeline}
        
\end{figure*}

\subsection{Enhancing Alignment Learning Pipeline}

Before delving into the details of our learning pipeline, we first clarify the necessary symbols and notations. We align the audio-text space of CLAP and the 3D-image space of ULIP (leaf-MCRs) to the image-text space of CLIP (base-MCR). The unimodal inputs of audio, text, image, and 3D point cloud are denoted as $A$, $T$, $V$, and $P$, respectively. The audio features, extracted from the CLAP audio encoder, are denoted as $\mathbf{A}^A = \{ \mathbf{a}_1^A, \mathbf{a}_2^A, ..., \mathbf{a}_{N}^A \}$. Similarly, employing the encoders of CLAP, CLIP, and ULIP, we can extract corresponding sets of features $\mathbf{T}^A$, $\mathbf{T}^I$, $\mathbf{V}^I$, $\mathbf{V}^U$, and $\mathbf{P}^U$, where the superscripts $A$, $I$, and $U$ represent encoding by CLAP, CLIP, and ULIP, respectively.

In Ex-MCR, freezing base-MCR allows us to preserve the original alignment of base-MCR but also implies that the modality gap within base-MCR remains preserved. Consequently, it becomes necessary to map the features of leaf-MCR to more suitable positions within the base-MCR space. To this end, we enhance the entire alignment learning pipeline from the perspectives of training data, architecture, and learning objectives. This enhanced learning pipeline effectively improves alignment's comprehensiveness, stability, and accuracy. Below, we sequentially introduce the design behind each perspective and corresponding motivations.

\subsubsection{Various Modality-centirc Data}
\label{Sec:data}
C-MCR employs data from overlapping modalities to aggregate semantic consistent embedding of non-overlapping modalities, thereby creating pseudo-pairs. This approach prompts a more comprehensive alignment. However, such a single modality-centric data is often biased and noisy. Taking CLIP and CLAP as an example, text encoders in CLIP and CLAP may introduce individual biases when encoding the same text, and texts can not describe the entire diverse visual or audio world. Therefore, the semantics of the text-centric pseudo-audio-image pairs are limited by the expressive power of text modality. The text-centric generated audio and image embeddings struggle to capture the entire audio and image representation space distribution.


To tackle the problem of limited and biased training data, we propose aggregating various modality-centric data. As depicted in the left sub-figure of Fig. \ref{fig:pipeline}, we no longer only take the overlapping modality as the query. Instead, all modalities in the two MCR spaces are iteratively employed as queries to aggregate corresponding semantically consistent embeddings.
Take aligning CLAP to CLIP as an example; the overlapping modality-centric (e.g., text-centric) consistent embeddings can be aggregated as follows: 
\begin{equation}
\begin{gathered}
    \label{eq:over-data}
    \tilde{\mathbf{t}}_i^A = \mathbf{t}_i^A; \ \ \ 
    \tilde{\mathbf{a}}_i^{A} = \operatorname{softmax}((\tilde{\mathbf{t}}_i^A \cdot \mathbf{T}^{A})/\tau_1) \cdot (\mathbf{T}^{A})^{T}; \\
    \tilde{\mathbf{t}}_i^I = \mathbf{t}_i^I; \ \ \ \tilde{\mathbf{v}}_i^{I} = \operatorname{softmax}((\tilde{\mathbf{t}}_i^I \cdot \mathbf{V}^{I})/\tau_1) \cdot (\mathbf{V}^{I})^{T}
\end{gathered}
\end{equation}
Where the $\tau_1$ is the temperature parameter of softmax. The $\tilde{\mathbf{t}}_i^A$ and $\tilde{\mathbf{t}}_i^I$ are derived from the same text data, and their semantics are natively consistent. Benefiting from the modality semantic alignment within each MCR, the generated $\tilde{\mathbf{a}}_i^{A}$ and $\tilde{\mathbf{v}}_i^{I}$ are also semantically relevant to the $\tilde{\mathbf{t}}_i^A$ and $\tilde{\mathbf{t}}_i^I$.

To capture the representation space of non-overlapping modality more comprehensively, we further introduce non-overlapping modality-centric (e.g., audio-centric or image-centric) data. This process (take audio-centric as an example) can be expressed as:
\begin{equation}
\begin{gathered}
    \label{eq:nonover-data}
    \tilde{\mathbf{a}}_i^A = \mathbf{a}_i^A; \ \ \ 
    \tilde{\mathbf{t}}_i^{A} = \operatorname{softmax}((\mathbf{a}_i^A \cdot \mathbf{T}^{A})/\tau_1) \cdot (\mathbf{T}^{A})^{T} \\
    \tilde{\mathbf{t}}_i^{I} = \operatorname{softmax}((\mathbf{a}_i^A \cdot \mathbf{T}^{A})/\tau_1) \cdot (\mathbf{T}^{I})^{T}; \ \ \ 
    \tilde{\mathbf{v}}_i^{I} = \operatorname{softmax}((\tilde{\mathbf{t}}_i^I \cdot \mathbf{V}^{I})/\tau_1) \cdot (\mathbf{V}^{I})^{T}
\end{gathered}
\end{equation}
Since the embeddings of $\mathbf{T}^A$ and $\mathbf{T}^I$ of overlapping modality are one-to-one matched, the similarity weights between $\mathbf{a}_i^A$ and $\mathbf{T}^{A}$ can be naturally transferred for aggregating embeddings of $\mathbf{T}^{I}$. These pseudo-embedding pairs derived from audio can better reflect the representation space of audio.
Based on the aforementioned formulas, when extending CLAP to CLIP, we can acquire three kinds (e.g., audio-centric, text-centric and image-centric) of semantically consistent embeddings $\{ \tilde{\mathbf{a}}_i^A, \tilde{\mathbf{t}}_i^{A}, \tilde{\mathbf{t}}_i^{I}, \tilde{\mathbf{v}}_i^{I} \}$. These three kinds of data are combined and shuffled for training.




\subsubsection{Decoupled Projector}
\label{sec:proj}
The main network structure of Ex-MCR is a projector, and it serves two purposes: 1) Learning the intra-MCR alignment to close the modality gaps within leaf-MCR and prompt more stable alignment between MCRs. 2) Learning the inter-MCR alignment for extending leaf-MCR to base-MCR. 
Considering these two different purposes, we propose a decoupled projector to alleviate the potential conflict between distinct optimization objectives while exploring a more reasonable mapping layer design for these two purposes. As shown in Fig. \ref{fig:pipeline}, the projector is decoupled into a linear layer $f_l(\cdot)$ for intra-MCR alignment and a multi-layer perceptron (MLP) layer $f_m(\cdot)$ for inter-MCR alignment. For the example of extending CLAP to CLIP, we first use $f_l$ to align $\tilde{\mathbf{a}}_i^A$ to $\tilde{\mathbf{t}}_i^A$ via L2 loss, which can be formulated as:
\begin{equation}
\label{l2}
    L_{intra} = \frac{1}{2} \frac{1}{B} \sum^B_{i=1}\lVert f_l(\tilde{\mathbf{a}}_i^{A})-\tilde{\mathbf{t}}_i^{A} \rVert_2
\end{equation}
With this intra-MCR alignment loss, $f_l(\cdot)$ learns the mapping between audio subspace and text subspace within the CLAP, thereby effectively closing the modality gap. Since the subspaces of different modalities in MCR space are actually very similar, linear mapping is enough to bridge the modality gap. Moreover, our experiments even found that activation layers have a negative effect on bridging the modality gap.

After bridging the modality gap, the shared $f_m(\cdot)$ are employed to map both audio and text embeddings of CLAP space to the CLIP space, which can be expressed as:
\begin{equation}
\label{f2}
    \hat{\mathbf{a}}^A_i = f_m(f_l(\tilde{\mathbf{a}}^A_i)); \ \ \hat{\mathbf{t}}^A_i=f_m(\mathbf{t}^A_i)
\end{equation}

\subsubsection{Dense Alignment Objective}
\label{sec:loss}
Since the modality gap within base-MCR is retained, a more robust learning objective is needed to map leaf-MCR to the appropriate position in the base-MCR space. To this end, we propose to learn the alignment densely among the quadruple semantic consistent pairs described in Sec. \ref{Sec:data}. In the case of the CLAP and CLIP, the dense inter-MCR alignment objectives are defined as:
\begin{equation}
\label{inter-MCR_loss}
    \begin{gathered}
        L_{avc} = \operatorname{InfoNCE}(\hat{\mathbf{a}}^{A},\tilde{\mathbf{v}}^{I});\ \ \ 
        L_{tvc} = \operatorname{InfoNCE}(\hat{\mathbf{t}}^{A},\tilde{\mathbf{v}}^{I})\\
        L_{atc} = \operatorname{InfoNCE}(\hat{\mathbf{a}}^{A},\tilde{\mathbf{t}}^{I});\ \ \ 
        L_{ttc} = \operatorname{InfoNCE}(\hat{\mathbf{t}}^{A},\tilde{\mathbf{t}}^{I})\\
    \end{gathered}
\end{equation}
where the $\operatorname{InfoNCE}(\cdot, \cdot)$ is the standard InfoNCE loss, which is defined as:
\begin{equation}
    \operatorname{InfoNCE}(\mathbf{x},\mathbf{z}) = - \frac{1}{2} \frac{1}{N} \sum_{i=1}^{N} \left [ \operatorname{log} \frac{\operatorname{exp}(\operatorname{sim}(\mathbf{x}_i, \mathbf{z}_i)/\tau_2)}{\sum_{j=1}^{N} \operatorname{exp}(\operatorname{sim}(\mathbf{x}_i, \mathbf{z}_j)/\tau_2)} + \operatorname{log} \frac{\operatorname{exp}(\operatorname{sim}(\mathbf{z}_i, \mathbf{x}_j)/\tau_2)}{\sum_{j=1}^{N} \operatorname{exp}(\operatorname{sim}(\mathbf{z}_i, \mathbf{x}_j)/\tau_2)} \right ]
    \label{eq:info}
\end{equation}
where the $\tau_2$ is the temperature parameter. The overall loss for extending CLAP to CLIP is defined as a weighted combination of the intra-MCR and inter-MCR losses:
\begin{equation}
\label{eq:loss}
    L = \lambda L_{intra} + \frac{1}{4}(L_{avc} + L_{atc} + L_{tvc} + L_{ttc})
\end{equation}
where $\lambda$ is the hyper-parameter to balance the two terms.

For ULIP and CLIP, symmetric various modality-centric data \ref{Sec:data}, decoupled projector \ref{sec:proj}, and dense alignment loss \ref{sec:loss} are employed to extend the 3D-image space to image-text space via the overlapping image modality.

Finally, we can use a unified representation space learning from existing MCRs for inference. Considering audio, text, image, and 3D point cloud inputs, we use CLAP's audio encoder, CLIP's text and image encoder, and ULIP's 3D encoder to extract the corresponding features $\mathbf{a}^A_i$, $\mathbf{t}^I_i$, $\mathbf{v}^I_i$, $\mathbf{p}^U_i$. And the $\mathbf{t}^I_i$, $\mathbf{v}^I_i$, $f_m^A(f_l^A(\mathbf{a}^A_i))$ and $f_m^U(f_l^U(\mathbf{p}^U_i))$ are the final audio-text-image-3D unified representation learned by Ex-MCR, where the $f_m^A(\cdot),f_l^A(\cdot); f_m^U(\cdot),f_l^U(\cdot)$ are the learned projectors of CLAP and ULIP respectively.

\section{Experiment}

\subsection{Experimental Setting}
\paragraph{Datasets}
\label{Sec:train_data}
For a fair comparison, we use the same unimodal datasets to C-MCR~\citep{wang2023connecting} for training, totaling 2.3M texts, 1.3M images, 1.8M audio, and 0.8M 3D point cloud. More details about training datasets are provided in the Appendix.

\paragraph{Implementation Details}
We employ pre-trained frozen CLIP ViT-B/32~\citep{radford2021learning}, CLAP~\citep{wu2023large}, and ULIP v2 (PointBERT version)~\citep{xue2023ulip2} models. The temperature $\tau_1$ in Eq. \ref{eq:over-data} \ref{eq:nonover-data} for embedding aggregation is set to 0.01 following \cite{wang2023connecting}, while the $\tau_2$ in \ref{eq:info} for InfoNCE loss calculation is set to 0.05. The hyper-parameter $\lambda$ in Eq. \ref{eq:loss} is set to 0.1. Following \cite{wang2023connecting}, we also add Gaussian noise with a variance of 0.004 to the semantic consistent embeddings described in Sec. \ref{Sec:data}. The linear projector $f_l(\cdot)$ is a simple linear layer, and the MLP projector $f_m(\cdot)$ is a 2-layer MLP. We train our model with a batch size of 4096 for 36 epochs. We employ the AdamW optimizer with an initial learning rate of 1e-3 and a cosine learning rate decay strategy.

\subsection{Audio-Visual-Text Experiment}
\paragraph{Downstream tasks.} 
We employ zero-shot audio-image, audio-text, and image-text retrieval tasks to evaluate the audio-image-text representations of Ex-MCR by extending CLAP to CLIP. For audio-image retrieval, we conduct evaluations on Flickr-SoundNet~\citep{senocak2018learning}, VGGSS~\citep{chen2021localizing}, and AVE~\citep{tian2018audio} datasets. Due to their small dataset sizes, we utilize all their available data, comprising 5,000, 5,000, and 4,097 samples. For audio-text retrieval, we utilize the validation set from the AudioCaps~\citep{kim2019audiocaps} dataset, which includes 964 audio samples, and for each audio, we choose one corresponding caption for retrieval. Regarding image-text retrieval, we employ the validation set of COCO~\citep{lin2014microsoft} dataset, consisting of 5,000 images, each accompanied by text captions. We randomly select one text annotation for each image as the ground truth. We calculate the cosine similarity between modalities in representation space and use mAP and Top-5 metrics for performance comparison.

\paragraph{Performance Comparison.}
Fig. \ref{CLAP_CLIP} compares Ex-MCR with WAV2CLIP, AudioCLIP, and C-MCR. Notably, even without using audio-image paired data, Ex-MCR achieves significantly better performance over WAV2CLIP and AudioCLIP, which illustrates that Ex-MCR is a more effective representation learning method when high-quality data pairs are limited. Furthermore, compared to C-MCR, Ex-MCR not only attains better audio-image alignment but also inherits more audio-text alignment from CLAP, with fully preserved image-text modality alignment of CLIP, which demonstrates the overall superiority of Ex-MCR over C-MCR in establishing new spaces and maintaining original spaces. In summary, extending CLAP to CLIP with our Ex-MCR method derives state-of-the-art audio-image-text unified representations.



\begin{table}[t]
\vspace{-2\baselineskip}
\caption{Results of audio-visual-text experiments. The best results are \textbf{bolded}. }
\centering
\label{CLAP_CLIP}
\renewcommand{\arraystretch}{1.1}
\resizebox{0.9\textwidth}{!}{
\begin{tabular}{@{}l|cccccc|cc|cc@{}}
\toprule
\multicolumn{1}{c|}{} & \multicolumn{6}{c|}{Audio-Image} & \multicolumn{2}{c|}{Audio-Text} & \multicolumn{2}{c}{Image-Text}  \\
\multicolumn{1}{l|}{\multirow{-2}{*}{Method}} & \multicolumn{2}{c}{FlickrNet} & \multicolumn{2}{c}{AVE}  & \multicolumn{2}{c|}{VGGSS}         & \multicolumn{2}{c|}{AudioCaps} & \multicolumn{2}{c}{COCO} \\ \midrule
                                              & mAP              & R@5              & mAP           & R@5           & mAP                                  & R@5                                   & mAP                                   & R@5                                   & mAP                                   & R@5                                   \\
{\color[HTML]{D8D8D8}CLAP}                                          & {\color[HTML]{D8D8D8}-}                & {\color[HTML]{D8D8D8}-}                & {\color[HTML]{D8D8D8}-}             & {\color[HTML]{D8D8D8}-}             & {\color[HTML]{D8D8D8}-}                                    & {\color[HTML]{D8D8D8}-}                                     & {\color[HTML]{D8D8D8} 21.98} & {\color[HTML]{D8D8D8} 35.23} & {\color[HTML]{D8D8D8}-}                                     & {\color[HTML]{D8D8D8}-}                                     \\
{\color[HTML]{D8D8D8}CLIP}                                          & {\color[HTML]{D8D8D8}-}                & {\color[HTML]{D8D8D8}-}                & {\color[HTML]{D8D8D8}-}             & {\color[HTML]{D8D8D8}-}             & {\color[HTML]{D8D8D8}-}                                    & {\color[HTML]{D8D8D8}-}                                     & {\color[HTML]{D8D8D8}-}                                     & {\color[HTML]{D8D8D8}-}                                     & {\color[HTML]{D8D8D8} 44.57} & {\color[HTML]{D8D8D8} 57.62} \\
AudioCLIP                                     & 3.81             & 4.91             & 2.33          & 2.65          & 3.10                                 & 3.94                                  & 2.23                                  & 2.68                                  & 20.14                                 & 27.42                                 \\
WAV2CLIP                                      & 2.77             & 3.41             & 3.48          & 4.23          & {\color[HTML]{D8D8D8} 7.42} & {\color[HTML]{D8D8D8} 10.47} & 0.88                                  & 0.99                                  & {44.57}                        & {57.62}                        \\
C-MCR                                         & {4.74}       & \textbf{5.97}    & {4.21}    & {4.91}    & {5.95}                                 & {7.69}                                  & {9.50}                                  & {13.62}                                 & 24.56                         & 33.83                           \\ \midrule
Ex-MCR                                        & \textbf{4.94}    & {5.95}       & \textbf{4.46} & \textbf{4.93} &  \textbf{6.39}                           & \textbf{8.12}                            & \textbf{11.19}                           & \textbf{16.65}                        & \textbf{44.57}                        & \textbf{57.62}                        \\ \bottomrule
\end{tabular}}
\end{table}

\subsection{3D-Visual-Text Results}
\begin{table}[t]
\centering
\label{ULIP_CLIP}
\vspace{-1\baselineskip}
\caption{Results of 3d-visual-text experiments.}
\setlength\tabcolsep{7pt}
\renewcommand{\arraystretch}{1.1}
\resizebox{0.9\textwidth}{!}{
\begin{tabular}{@{}l|ccc|ccc|ccc@{}}
\toprule
\multicolumn{1}{l|}{\multirow{2}{*}{Method}} & \multicolumn{3}{c|}{3D-Text}                     & \multicolumn{3}{c|}{3D-Image}                  & \multicolumn{3}{c}{Image-Text}              \\
\multicolumn{1}{c|}{}                        & \multicolumn{3}{c|}{ModelNet40}                  & \multicolumn{3}{c|}{Objaverse-LVIS}             & \multicolumn{3}{c}{COCO}                         \\ \midrule
                                             & Acc@1           & Acc@3           & Acc@5           & mAP            & R@1           & R@5            & mAP            & R@1            & R@5            \\
{\color[HTML]{D8D8D8}CLIP}                                         & {\color[HTML]{D8D8D8}-}          & {\color[HTML]{D8D8D8}-}  & {\color[HTML]{D8D8D8}-}          & {\color[HTML]{D8D8D8}-}           & {\color[HTML]{D8D8D8}-}   & {\color[HTML]{D8D8D8}-}  &  {\color[HTML]{D8D8D8}44.57}  &  {\color[HTML]{D8D8D8}32.58}   & {\color[HTML]{D8D8D8}57.62}  \\
ULIP                                         & 60.40          & 79.00          & 84.40          & 3.54           & 1.45          & 4.51           & {34.42}    & {22.92}    & {46.33}    \\
ULIP v2                                       & \textbf{73.06} & 86.39          & 91.50          & \textbf{11.41} & \textbf{6.00} & \textbf{15.63} & {34.42}    & {22.92}    & {46.33}    \\
C-MCR                                        & 64.90          & {87.00}    & {92.80}    & 3.84           & 1.36          & 4.80           & 24.23          & 14.34          & 33.19          \\ \midrule
Ex-MCR                                       & {66.53}    & \textbf{87.88} & \textbf{93.60} & {6.23}     & {2.54}    & {8.25}     & \textbf{44.57} & \textbf{32.58} & \textbf{57.62} \\ \bottomrule
\end{tabular}}
\end{table}

\paragraph{Downstream tasks.}
To evaluate the performance of 3D-image-text space learned by extending ULIP to CLIP, we conduct a zero-shot 3D object classification task to assess the alignment between 3D and text. We also perform zero-shot 3D-image and image-text retrieval tasks to evaluate the alignment between 3D and image, as well as image and text.
The zero-shot 3D object classification task is carried on the ModelNet40~\citep{wu20153d} validation set, comprising 2468 sample pairs across 40 different classes. We embed the label into 64 prompt templates for each class, then extracted and averaged the features to obtain the corresponding text embeddings, following~\cite{xue2023ulip2}.
Regarding the zero-shot 3D-image retrieval task, we use the Objaverse-LVIS dataset~\citep{deitke2023objaverse}, which includes 46,054 3D objects. For each 3D object, ULIP v2 provides 12 rendered images from different perspectives, and we randomly selected one as the paired image. Additionally, we continued to use the COCO dataset's validation set for zero-shot image-text retrieval.

\paragraph{Performance Comparison}
From Tab. \ref{ULIP_CLIP}, we can find the following key points. Firstly, even without using any 3D-text data, Ex-MCR still outperforms the advanced models (ULIP and ULIP v2) trained on 3D-text pairs in most performance metrics for 3D object classification. Secondly, the 3D-image retrieval accuracy of Ex-MCR is significantly higher than ULIP and C-MCR but lower than ULIP v2. Since the 3D-image space of ULIP v2 is treated as leaf-MCR, it is reasonable that Ex-MCR 3D-image performance is slightly lower than ULIP v2. At the same time, the better 3D-image retrieval accuracy than ULIP and C-MCR shows that Ex-MCR effectively learns strong 3D-image alignment.
Finally, Ex-MCR retains the best image-text retrieval accuracy compared to these previous state-of-the-art models. The leading performance on all these tasks further demonstrates the superiority of Ex-MCR in unified contrastive representation learning.

\subsection{Emergent 3D-Audio Alignment}
\begin{figure}[t]
\centering
\label{audio2pcd}
\includegraphics[width=0.95\textwidth]{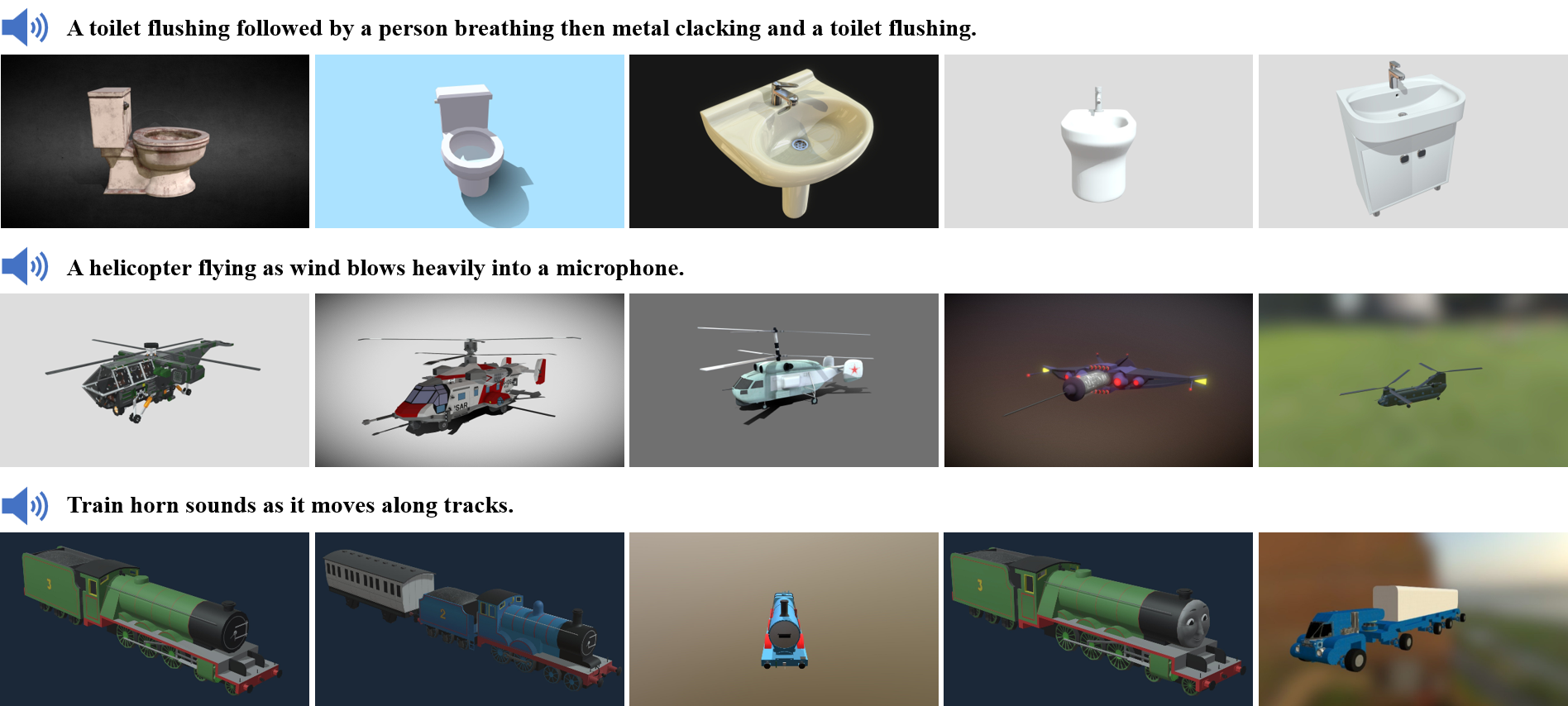}
\vspace{-1\baselineskip}
\caption{Visualization of Audio to 3D retrieval.}
\vspace{-1\baselineskip}
\label{audio2pcd}
\end{figure}

\begin{figure}[t]
\centering
\includegraphics[width=0.95\textwidth]{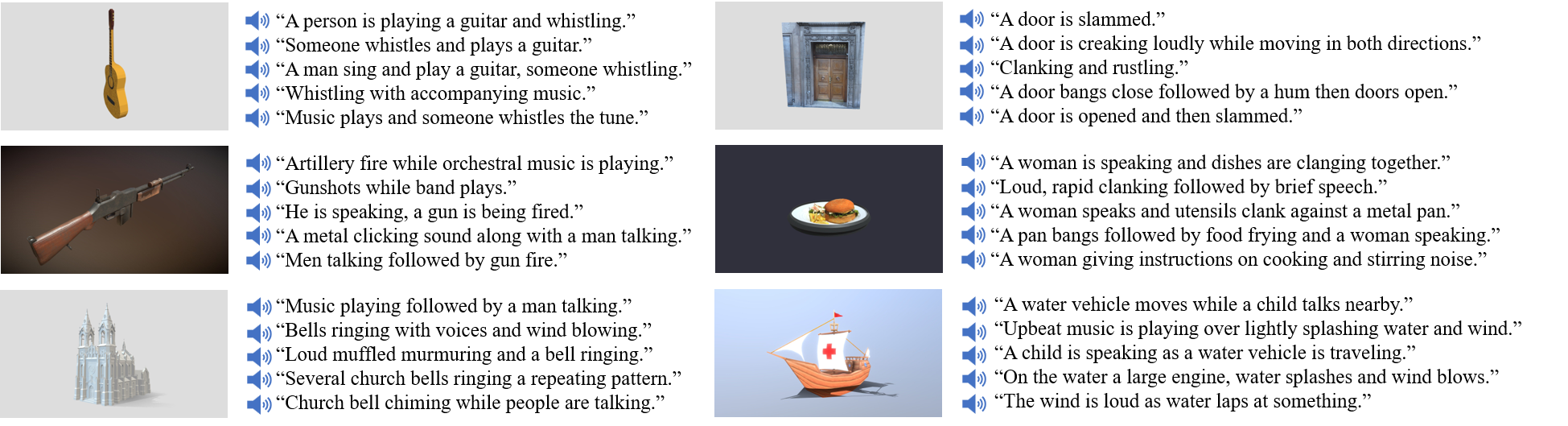}
\vspace{-1\baselineskip}
\caption{Visualization of 3D to Audio retrieval.}
\vspace{-1\baselineskip}
\label{pcd2audio}
\end{figure}
In this section, we study whether the semantic alignment also emerges between the extended modalities (e.g., audio and 3D). We mutually retrieve audio in AudioCaps and 3D objects in Objaverse. In Fig.~\ref{audio2pcd} and \ref{pcd2audio}, we provide visualizations of some top-5 retrieval results, and audios are described by their corresponding caption annotations.
These cases effectively demonstrate the emergent semantic alignment between audio-3D in Ex-MCR space.
For example, the sound of a flushing toilet and water flow can retrieve 3D objects of toilets or sinks, while a sailboat 3D object can retrieve clips containing sounds of water vessels and wind. More results and the original audio files are provided in our supplementary material.

These exciting results demonstrate that extending ULIP and CLAP onto CLIP following our Ex-MCR methods derives a 3D-vision-text-audio unified contrastive representation space. In addition to the state-of-the-art performance on all possible tasks, Ex-MCR is an extremely training-efficient and paired-data-free representation learning method, which amplifies its application value in unified multi-modal representation learning.

\subsection{Ablation Studies}

In this section, we analyze the main components of Ex-MCR. All experiments are conducted on extending CLAP to CLIP, and we reported the average mAP of audio-visual and audio-text retrieval on AVE and AudioCaps datasets, respectively. In addition, we also provide results on more datasets and evaluation metrics in the Appendix.

\begin{minipage}[t]{\textwidth}
\begin{minipage}[t]{0.33\textwidth}
\makeatletter\def\@captype{table}
\setlength\tabcolsep{9pt}
\renewcommand{\arraystretch}{1.2}
\caption{Data modality-centric. A, I, and T represent audio-centric, image-centric, and text-centric data, respectively.}
\label{tab:1}
\begin{tabular}{@{}c|cc@{}}
\toprule
      & AVE           & AudioCaps      \\ \midrule
A     & 4.10          & 11.11 \\
I     & 3.41          & 5.54           \\
T     & 4.17 & 9.89           \\
A+I   & 4.11          & 11.09          \\
A+T   & 4.12          & 10.88          \\
I+T   & 4.05          & 8.39           \\ \midrule
A+I+T & \textbf{4.46}          & \textbf{11.19}          \\ \bottomrule
\end{tabular}
\end{minipage}
\hspace{0.1cm}
\begin{minipage}[t]{0.3\textwidth}
\makeatletter\def\@captype{table}
\setlength\tabcolsep{5pt}
\renewcommand{\arraystretch}{1.28}
\centering
\caption{Alignment objective. A-T, T-T, A-V, and T-V represent the alignment objective between audio-text, text-text, audio-image, and text-image, respectively.}
\vspace{0.5\baselineskip}
\label{tab:2}
\begin{tabular}{@{}c|cc@{}}
\toprule
                  & AVE           & AudioCaps      \\ \midrule
A-T                & 4.00          & 10.82          \\
T-T                & 4.15          & 11.30 \\
A-V                & 3.97          & 7.49           \\
T-V                & 4.18 & 7.68           \\ \midrule
All             & \textbf{4.46}          & \textbf{11.19}          \\ \bottomrule
\end{tabular}
\end{minipage}
\hspace{0.1cm}
\begin{minipage}[t]{0.33\textwidth}
\makeatletter\def\@captype{table}
\setlength\tabcolsep{7pt}
\renewcommand{\arraystretch}{1.1}
\caption{Structure of $f_1(\cdot)$}
\label{tab:3}
\begin{tabular}{@{}c|cc@{}}
\toprule
$f_1(\cdot)$ & AVE           & AudioCaps      \\ \midrule
Linear           & \textbf{4.46} & \textbf{11.19} \\
1 MLP            & 4.16          & 10.25          \\
2 MLP            & 4.04          & 9.93           \\ \bottomrule
\end{tabular}
\setlength\tabcolsep{7pt}
\renewcommand{\arraystretch}{1.1}
\vspace{-0.5\baselineskip}
\caption{Structure of $f_m(\cdot)$}
\label{tab:4}
\begin{tabular}{@{}c|cc@{}}
\toprule
$f_m(\cdot)$ & AVE           & AudioCaps      \\ \midrule
Linear           & 3.70          & 11.15          \\
1 MLP            & 4.15          & 10.53          \\
2 MLP            & \textbf{4.46} & 11.19          \\
3 MLP            & 4.31          & \textbf{11.30} \\
4 MLP            & 4.35          & 11.07          \\
5 MLP            & 4.42          & 10.93          \\ \bottomrule
\end{tabular}
\label{sample-table}
\end{minipage}
\end{minipage}

\paragraph{Various modality-centric data} As described in Sec. \ref{Sec:data}, We employ various modality-centric data to train our projectors. For investigating the effect of different modality-centric data, we ablate each modality-centric data, and the results are reported in Tab. \ref{tab:1}. Each kind of data is beneficial for audio-visual and audio-image alignment, and using all kinds of data simultaneously brings the best performance.

\paragraph{Dense alignment objective}
To analyze the impact of different alignment objectives, we train the model with each alignment objective. From the results reported in Tab. \ref{tab:2}, we can find that aligning the overlapping modalities (text) is most important. In terms of audio-visual alignment, directly aligning the pseudo-audio-visual pairs shows sub-optimal performance, which proves that the pseudo-data aggregating process is biased and noisy, and the pseudo-audio-visual data pair is most severely affected by noise.
On the other hand, for audio-text alignment, the model using text-text alignment objective surpasses that of directly aligning pseudo-audio-text pairs, which further demonstrates the importance of alignment learned from overlapping modality.

\paragraph{Structure of $f_l(\cdot)$}
Tab. \ref{tab:3} demonstrates the impact of different structures of $f_l(\cdot)$. The results prove our hypothesis: the representation structures between different modalities within one MCR space are similar, and a simple linear layer is enough to bridge the modality gap. Moreover, MLP with an activation layer introduces non-linearity, which may disrupt the representation's spatial structure, bringing sub-optimal performance.

\paragraph{Structure of $f_m(\cdot)$}
The impact of structures of $f_m(\cdot)$ is summarized in Tab. \ref{tab:4}. For aligning to distinct MCR space, the non-linear MLP structure is better than the simple linear layer. Besides, our experiments show that 2-layer MLP may be enough, and more layers would not bring further performance improvement. 

\section{Conclusion}
This paper proposes \textbf{Ex}tending \textbf{M}ulti-modal \textbf{C}ontrastive \textbf{R}epresentations (Ex-MCR), a novel training-efficient and paired-data-free unified constrastive representation learning method for more than three modalities. Ex-MCR effectively integrates the knowledge in pre-trained MCRs through overlapping modalities between these MCRs. By extending ULIP and CLAP onto CLIP via the overlapping image and text modality, respectively, we derive unified and high-quality audio-image-text-3D representations. Without using any paired data, Ex-MCR attains a series of state-of-the-art performance results across various tasks. More importantly, semantic alignment is also observed between extended modalities (e.g., audio-3D), which highlights the potential of Ex-MCR in modality extensibility.

\bibliography{iclr2024_conference}
\bibliographystyle{iclr2024_conference}

\newpage

\appendix
\section{Training Dataset}
The details of our training dataset, which are mentioned in Sec. \ref{Sec:train_data}, are shown below.
\paragraph{Text Dataset.}
To ensure that the texts contain sufficient information for other modalities, the data of text is sourced from diverse perspectives in vision-text datasets (COCO, CC3M), video-text datasets (MSRVTT, MAD), and audio-text datasets (AudioCaps, Clotho). Following~\cite{wang2023connecting}, we select 1M texts from CC3M. There are 2.33M text samples in total. We extract their CLAP and CLIP features $\mathbf{T}^A$ and $\mathbf{T}^I$ using the CLAP and CLIP encoders, respectively.

\paragraph{Image Dataset.}
For another modality in base-MCR, Vision, we utilize ImageNet1K as the data source. ImageNet1K is a large-scale image recognition dataset consisting of 1.3 million images. We extract their features to the sets $\mathbf{V}^I$, and $\mathbf{V}^U$ in CLIP and ULIP, using the CLIP Encoder and ULIP Encoder.

\paragraph{Audio Dataset.}
AudioSet is a large-scale audio dataset with 2.1M audio clips from YouTube, equivalent to 5.8 thousand hours of audio and encompassing over 500 sound classes. 
We use the CLAP audio encoder to extract the feature set $\mathbf{A}^A$ from the audios of the training set.


\paragraph{3D Point Cloud Dataset.}
For the 3D modality, we use Objaverse, the recently released and large-scale 3D objects dataset. It has approximately 800K real-world 3D objects. All 3D data are transformed into point clouds and extracted into the feature set $\mathbf{P}^U$ using the ULIP 3D encoder.

It is worth noting that we do not employ any annotations provided with the datasets mentioned above as part of our training data, which means we only use the unimodal modality of data in each dataset we selected.

\section{Architecture of Projectors}

\begin{table}[h]
\centering
\caption{Model configurations of projectors.}
\label{tab:model}
\begin{tabular}{@{}c|c|cc@{}}
\toprule
Module       & Block       & $C_{in}$ & $C_{out}$ \\ \midrule
$f_1(\cdot)$ & Linear      & 512      & 512       \\ \midrule
\multirow{12}{*}{$f_m(\cdot)$} & Linear      & 512      & 1024      \\
             & BatchNorm1D & 1024     & 1024      \\
             & Relu        & -        & -         \\
             & Linear      & 1024     & 512       \\
             & BatchNorm1D & 512      & 512       \\
             & Relu        & -        & -         \\
             & Linear      & 512      & 1024      \\
             & BatchNorm1D & 1024     & 1024      \\
             & Relu        & -        & -         \\
             & Linear      & 1024     & 512       \\
             & BatchNorm1D & 512      & 512       \\
             & Relu        & -        & -         \\\bottomrule
\end{tabular}
\end{table}

The model configurations of our projectors are shown in Tab. \ref{tab:model}.

\section{Detailed Results of Ablation Study}
As a supplement to Tab. \ref{tab:1}, Tab. \ref{tab:2}, Tab. \ref{tab:3}, and Tab. \ref{tab:4}, we provide detailed ablation experiment results on more comprehensive evaluation metrics of various datasets, as shown below.


\begin{table}[h]
\centering
\caption{Detailed results of experiments on data modality-centric.}
\begin{tabular}{@{}c|cc|cc|cc|cc@{}}
\toprule
Data Perspective & \multicolumn{2}{c|}{FlickrNet} & \multicolumn{2}{c|}{AVE} & \multicolumn{2}{c|}{VGGSS} & \multicolumn{2}{c}{AudioCaps} \\ \midrule
       & mAP            & R@5           & mAP         & R@5        & mAP          & R@5         & mAP           & R@5           \\
A      & 3.94           & 4.77          & 4.10        & 4.66       & 5.47         & 6.95        & 11.11         & 16.39         \\
I      & 3.83           & 4.63          & 3.41        & 3.70       & 4.82         & 5.96        & 5.54          & 7.18          \\
T      & 4.85           & 5.96          & 4.17        & 4.61       & 5.72         & 7.23        & 9.89          & 14.47         \\
A+I    & 4.22           & 4.96          & 4.11        & 4.71       & 6.01         & 7.78        & 11.09         & 16.91         \\
A+T    & 4.63           & 5.56          & 4.12        & 4.64       & 5.88         & 7.57        & 10.88         & 16.23         \\
I+T    & 4.70           & 5.82          & 4.05        & 4.34       & 5.84         & 7.36        & 8.39          & 12.09         \\
A+I+T  & 4.94           & 5.95          & 4.46        & 4.93       & 6.39         & 8.12        & 11.19         & 16.65         \\ \bottomrule
\end{tabular}
\end{table}

\begin{table}[h]
\centering
\caption{Detailed results of experiments on the structure of $f_1(\cdot)$.}
\begin{tabular}{@{}c|cc|cc|cc|cc@{}}
\toprule
$f_1(\cdot)$ & \multicolumn{2}{c|}{FlickrNet} & \multicolumn{2}{c|}{AVE} & \multicolumn{2}{c|}{VGGSS} & \multicolumn{2}{c}{AudioCaps} \\ \midrule
       & mAP            & R@5           & mAP         & R@5        & mAP          & R@5         & mAP           & R@5           \\
Linear & 4.94           & 5.95          & 4.46        & 4.93       & 6.39         & 8.12        & 11.19         & 16.65         \\
1 MLP  & 4.54           & 5.59          & 4.16        & 4.75       & 6.50         & 8.54        & 10.25         & 14.92         \\
2 MLP  & 4.36           & 5.15          & 4.04        & 4.66       & 6.00         & 7.63        & 9.93          & 14.48         \\ \bottomrule
\end{tabular}
\end{table}

\begin{table}[h]
\centering
\caption{Detailed results of experiments on the structure of $f_m(\cdot)$.}
\begin{tabular}{@{}c|cc|cc|cc|cc@{}}
\toprule
$f_m(\cdot)$ & \multicolumn{2}{c|}{FlickrNet} & \multicolumn{2}{c|}{AVE} & \multicolumn{2}{c|}{VGGSS} & \multicolumn{2}{c}{AudioCaps} \\ \midrule
       & mAP            & R@5           & mAP         & R@5        & mAP          & R@5         & mAP           & R@5           \\
Linear & 3.62           & 4.50          & 3.70        & 4.03       & 5.40         & 6.82        & 11.15         & 16.37         \\
1 MLP  & 4.62           & 5.79          & 4.15        & 4.76       & 5.81         & 7.28        & 10.53         & 15.87         \\
2 MLP  & 4.94           & 5.95          & 4.46        & 4.93       & 6.39         & 8.12        & 11.19         & 16.65         \\
3 MLP  & 4.85           & 5.93          & 4.31        & 4.88       & 6.57         & 8.70        & 11.30         & 17.10         \\
4 MLP  & 4.95           & 6.20          & 4.35        & 4.84       & 6.55         & 8.57        & 11.07         & 16.23         \\
5 MLP  & 4.79           & 6.02          & 4.42        & 5.15       & 6.59         & 8.63        & 10.93         & 16.21         \\ \bottomrule
\end{tabular}
\end{table}

\begin{table}[h]
\centering
\caption{Detailed results of experiments on alignment objective.}
\begin{tabular}{@{}c|cc|cc|cc|cc@{}}
\toprule
Objective & \multicolumn{2}{c|}{FlickrNet} & \multicolumn{2}{c|}{AVE} & \multicolumn{2}{c|}{VGGSS} & \multicolumn{2}{c}{AudioCaps} \\ \midrule
       & mAP            & R@5           & mAP         & R@5        & mAP          & R@5         & mAP           & R@5           \\
A-T    & 4.01           & 4.78          & 4.00        & 4.56       & 5.70         & 7.28        & 10.82         & 15.87         \\
T-T    & 4.56           & 5.33          & 4.15        & 4.54       & 5.68         & 6.86        & 11.30         & 16.93         \\
A-V    & 4.30           & 5.34          & 3.97        & 4.51       & 5.91         & 7.30        & 7.49          & 10.35         \\
T-V    & 4.77           & 6.03          & 4.18        & 4.92       & 5.43         & 6.93        & 7.68          & 10.36         \\
Dense  & 4.94           & 5.95          & 4.46        & 4.93       & 6.39         & 8.12        & 11.19         & 16.65         \\ \bottomrule
\end{tabular}
\end{table}

\end{document}